# Stochastic Variational Inference with Gradient Linearization

Tobias Plötz*   Anne S. Wannenwetsch*   Stefan Roth
Department of Computer Science, TU Darmstadt

## Abstract

*Variational inference has experienced a recent surge in popularity owing to stochastic approaches, which have yielded practical tools for a wide range of model classes. A key benefit is that stochastic variational inference obviates the tedious process of deriving analytical expressions for closed-form variable updates. Instead, one simply needs to derive the gradient of the log-posterior, which is often much easier. Yet for certain model classes, the log-posterior itself is difficult to optimize using standard gradient techniques. One such example are random field models, where optimization based on gradient linearization has proven popular, since it speeds up convergence significantly and can avoid poor local optima. In this paper we propose stochastic variational inference with gradient linearization (SVIGL). It is similarly convenient as standard stochastic variational inference – all that is required is a local linearization of the energy gradient. Its benefit over stochastic variational inference with conventional gradient methods is a clear improvement in convergence speed, while yielding comparable or even better variational approximations in terms of KL divergence. We demonstrate the benefits of SVIGL in three applications: Optical flow estimation, Poisson-Gaussian denoising, and 3D surface reconstruction.*

## 1. Introduction

Computer vision algorithms increasingly become building blocks in ever more complex systems, prompting for ways of assessing the reliability of each component. Probability distributions allow for a natural way of quantifying predictive uncertainty. Here, variational inference (VI, see [43] for an extensive introduction) is one of the main computational workhorses. Stochastic approaches to variational inference [17, 21, 31, 33] have recently rejuvenated the interest in this family of approximate inference methods. Part of their popularity stems from their making variational inference applicable to large-scale models, thus enabling practical systems [40]. Another benefit, which should not be underestimated, is that they allow to apply variational inference in a black-box fashion [31, 40], since it is no longer required to carry out tedious and moreover model-specific derivations of the update equations. This allows practitioners to apply variational inference to new model classes very quickly. The only required model specifics are gradients of the log-posterior w.r.t. its unknowns, which are typically much easier to derive than variational update equations. Moreover, automatic differentiation [4] can be used to further reduce manual intervention.

While this makes stochastic variational inference techniques attractive from the user's perspective, there are some caveats. In this paper we specifically focus on the limitations of gradient-based optimization techniques in the context of certain highly nonlinear model classes. One such category are random field models [6], which often arise in dense prediction tasks in vision. Let us take optical flow [8, 32] as an illustrative example. The data model is highly multimodal and the prior frequently relies on non-convex potentials, which complicate inference [5]. Gradient-based optimization is severely challenged by the multi-modal energy function. Hence, approaches based on energy minimization [8, 32, 42] often rely on a optimization technique called gradient linearization [29], which proceeds by iteratively linearizing the gradient at the current estimate and then solving the resulting system of linear equations to obtain the next iterate. Our starting point is the following question: If gradient linearization is beneficial for maximum a-posteriori (MAP) estimation in certain model classes, would not stochastic variational inference benefit similarly?

In this paper, we derive *stochastic variational inference with gradient linearization (SVIGL)* – a general optimization algorithm for stochastic variational inference that only hinges on the availability of linearized gradients of the underlying energy function. In each iteration, SVIGL linearizes a stochastic gradient estimate of the Kullback-Leibler (KL) divergence and solves for the root of the linearization. We show that each step of this procedure optimizes a sound objective. Furthermore, we make interesting experimental findings for challenging models from optical flow estimation and Poisson-Gaussian denoising. First, we observe that SVIGL leads to faster convergence of the variational objective function than gradient-based stochas-tic variational inference (SVI) with the strong optimizers Adam [20] and stochastic gradient descent (SGD). Second, we show that SVIGL is more robust w.r.t. its optimization parameters than standard gradient-based approaches. Finally, SVIGL enables re-purposing existing well-proven energy minimization schemes and implementations to obtain uncertainty estimates while maintaining, or even improving, application performance. Figure 1 shows exemplary flow fields and uncertainty predictions of SVIGL. As expected intuitively, high uncertainty values coincide with errors in the estimated flow field, *e.g.* near motion discontinuities. Finally, we show that SVIGL benefits problems beyond dense prediction by employing it for 3D surface reconstruction.

---

*Authors contributed equally

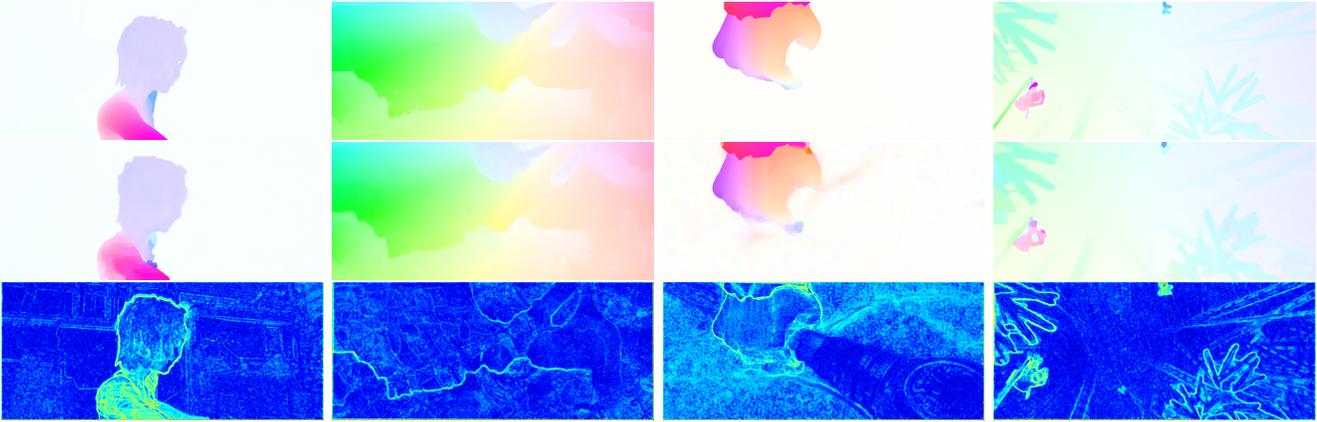

Figure 1. *Example application of SVIGL to optical flow estimation:* Ground truth (top), flow predictions (middle), and uncertainty estimates (bottom) on Sintel final [9]. Note that the uncertainties agree well with the flow errors.

## 2. Related Work

**Variational inference.** For Bayesian networks, VI w.r.t. to the exclusive Kullback-Leibler divergence KL $(q \| p)$ has usually been restricted to certain model classes. The parametric form of the approximating distribution $q$ is chosen such that update equations for the variational parameters of $q$ are analytically tractable. Here, conjugate-exponential models [47] are very common as they often arise in the context of topic modeling, *e.g.* in the LDA model [7, 38].

In other application areas, *e.g.* in computer vision, Markov random field (MRF) models are more common. Traditionally, VI has only been applied to specific model classes with closed-form updates, *e.g.* [11, 23, 24, 27, 36]. Miskin and MacKay [27] pioneered the use of VI for Bayesian blind deconvolution, but made the restrictive assumption that the prior is fully factorized. Levin *et al.* [23] use a mixture of Gaussian prior on the image derivatives. However, this more powerful prior comes at the cost of additionally maintaining a variational approximation of all mixture components. Krähenbühl and Koltun [22] consider fully-connected conditional random fields (CRF) with Gaussian edge potentials. In this special case mean-field inference can be done efficiently through filtering. Schelten and Roth [36] apply VI to high-order random fields.

In all of the previously mentioned works the variational inference algorithm is closely tied to the probabilistic model at hand and oftentimes requires tedious derivations of analytical update equations. In this paper, we aim to make VI more practical as the only interaction with the probabilistic model is through the linearized gradient of its log probability density function, thus allowing for easy variational inference for a rich class of graphical models.

**Stochastic variational optimization.** Recently, it was shown that the KL divergence is amenable to stochastic optimization if the approximating distribution $q$ can be reparameterized in terms of a base distribution that does not or only weakly depend on the parameters of $q$ [21, 33, 35]. While SVI was originally proposed for learning deep latent variable models, such as variational auto-encoders, it is also applicable more generally to graphical models. Reparameterization allows for deriving efficient stochastic estimators of the gradient of the KL divergence [21, 28]. Only the unnormalized log-density and its gradient w.r.t. the hidden variables are required, thus enabling black-box VI [19, 31, 40]. Note that by stochastic variational inference we do not just refer to the method of Hoffman *et al.* [17], which, in contrast, requires the true posterior to be from the conjugate-exponential family. Instead, we use the term more generally to describe VI using stochastic optimization.

Having access to a gradient estimator, stochastic algorithms [34] are employed to do the actual optimization. Nowadays, one of the default choices is Adam [20], but other approaches are in use as well, *e.g.* RMSprop [39], AdaGrad [13], or L-BFGS-SGVI [14]. These algorithms each implement a gradient descent method that is tuned with the recent history of gradient evaluations. In contrast, we assume that we observe a linearization of the gradient and use the information contained therein to modify the direction of the parameter updates. This can be seen as a gradient descent with a special preconditioner [29], see supplemental.

**Applications of uncertainties.** Aside from being a popular inference tool in many areas of computer vision, *e.g.* [22, 23], VI yields an assessment of the uncertainty, which can be exploited to post-process point estimates, *e.g.* with the fast bilateral solver [3]. When used as input for higher-level tasks, optical flow uncertainties allow to discard unreliable estimates and avoid error propagation [45], *e.g.* in image segmentation [30] or tracking [46]. Uncertainties in image restoration can be beneficial in video restoration, where estimates are fused over several frames [12].

## 3. Preliminaries

*Variational inference* [43] generally aims to approximate an intractable distribution $p$ with a tractable distribution $q$. Since our applications are based on CRFs, we will specifically look at finding approximations to a posterior distribution $p(\mathbf{x} \mid \mathbf{y})$. Note, however, that our approach can be applied to marginal and joint distributions as well. We assume that $p$ is a density function over continuous variables, and can be expressed as a Gibbs distribution with its energy function $E(\mathbf{x}, \mathbf{y})$ and partition function $Z(\mathbf{y})$ as

$$p(\mathbf{x} \mid \mathbf{y}) = \frac{1}{Z(\mathbf{y})} \exp \big\{ -E(\mathbf{x}, \mathbf{y}) \big\}. \tag{1}$$

To ease notation, we assume the temperature parameter to be subsumed into $E(\mathbf{x}, \mathbf{y})$, which we furthermore assume to be differentiable. The approximating distribution $q$ is chosen to be a member of some parameterized family of distributions with parameter $\boldsymbol{\theta}$, usually from the exponential family [43]. To determine $q$, variational inference then aims to find variational parameters $\hat{\boldsymbol{\theta}}$ that minimize the exclusive Kullback-Leibler divergence $\mathrm{KL}\,(q \,\|\, p)$, *i.e.*

$$\hat{\boldsymbol{\theta}} = \arg\min_{\boldsymbol{\theta}} \mathrm{KL}\,(q \,\|\, p) \tag{2a}$$

$$= \arg\min_{\boldsymbol{\theta}} -\mathbb{E}_{q(\mathbf{x};\boldsymbol{\theta})}[\log p(\mathbf{x} \mid \mathbf{y})] + \mathbb{E}_{q(\mathbf{x};\boldsymbol{\theta})}[\log q(\mathbf{x};\boldsymbol{\theta})] \tag{2b}$$

$$= \arg\min_{\boldsymbol{\theta}} -\mathbb{E}_{q(\mathbf{x};\boldsymbol{\theta})}[\log p(\mathbf{x} \mid \mathbf{y})] - H(q), \tag{2c}$$

where $H(q) = H\big(q(\mathbf{x};\boldsymbol{\theta})\big)$ denotes the entropy of $q$.

**Gradient linearization.** We now take a step back and first look at MAP estimation for the energy $E(\mathbf{x}, \mathbf{y})$ in Eq. (1), *i.e.* the problem of finding

$$\hat{\mathbf{x}} = \arg\max_{\mathbf{x}} \log p(\mathbf{x} \mid \mathbf{y}) = \arg\min_{\mathbf{x}} E(\mathbf{x}, \mathbf{y}). \tag{3}$$

Assuming that $E$ is differentiable, we could now apply a standard gradient method, but this may lead to slow convergence. On the other hand, second-order methods may be difficult to apply as the Hessian can be tedious to obtain and/or too dense. For many large-scale prediction problems in computer vision, *e.g.* estimating optical flow [8, 32], denoising [41], or deblurring [42], this has been addressed through iterative *gradient linearization* (GL). In this procedure, given a current estimate $\mathbf{x}^{(t)}$, the gradient of the energy function $E$ w.r.t. $\mathbf{x}$ is linearized around $\mathbf{x}^{(t)}$ as

$$\nabla_{\mathbf{x}} E(\mathbf{x}, \mathbf{y}) \approx \bar{\nabla}_{\mathbf{x}} E\big(\mathbf{x}; \mathbf{x}^{(t)}\big) = \mathbf{A}_{\mathbf{x}}\big(\mathbf{x}^{(t)}\big)\mathbf{x} + \mathbf{b}_{\mathbf{x}}\big(\mathbf{x}^{(t)}\big). \tag{4}$$

For notational brevity, we omit $\mathbf{y}$ here and in the following. Note that the linearized gradient $\bar{\nabla}_{\mathbf{x}} E\big(\mathbf{x}; \mathbf{x}^{(t)}\big)$ is exact at $\mathbf{x} = \mathbf{x}^{(t)}$. To obtain the next iterate $\mathbf{x}^{(t+1)}$, we set $\bar{\nabla}_{\mathbf{x}} E$ to zero and solve the resulting linear system of equations

$$\mathbf{x}^{(t+1)} = -\mathbf{A}_{\mathbf{x}}^{-1}\big(\mathbf{x}^{(t)}\big)\mathbf{b}_{\mathbf{x}}\big(\mathbf{x}^{(t)}\big) \tag{5}$$

using an exact or approximate standard solver. Like in any iterative optimization, an initial guess $\mathbf{x}^{(0)}$ is required.

Iterative GL is also known by various other names. Nikolova and Chan [29] showed it to be equivalent to multiplicative half-quadratic minimization [16] for Gaussian likelihoods. Moreover, it is closely related to iteratively reweighted least squares through their equivalence to half-quadratic approaches [18]. Finally, GL can be seen as preconditioned gradient descent using $\mathbf{A}_{\mathbf{x}}^{-1}$ as preconditioner [29], *c.f.* supplemental. In comparison to Newton's method no second-order derivatives are required – a benefit that is shared with other quasi-Newton methods, such as the popular L-BFGS [10]. However, every regular gradient step couples variables only within a local spatial neighborhood. In contrast, one iteration of GL (Eq. 5) causes a joint update of all variables leading to faster convergence in highly non-linear objectives (see Fig. 2 for an example).

## 4. Stochastic Variational Inference with Gradient Linearization (SVIGL)

We now aim to leverage the advantages of GL in the context of stochastic variational inference. To that end, we assume access to a linearized gradient, given by $\mathbf{A}_{\mathbf{x}}$ and $\mathbf{b}_{\mathbf{x}}$ in Eq. (4). By applying the re-parameterization trick [21, 33], we can rewrite the KL divergence of Eq. (2) as

$$\hat{\boldsymbol{\theta}} = \arg\min_{\boldsymbol{\theta}} -\mathbb{E}_{\mathbf{z} \sim \mathcal{G}}\Big[ \log p\big(\mathbf{x}(\mathbf{z}) \mid \mathbf{y}\big) \Big] - H(q), \tag{6}$$

where $\mathbf{x}(\mathbf{z}) \equiv \mathbf{x}(\mathbf{z}; \boldsymbol{\theta})$, and $\mathbf{z}$ is distributed following a base distribution $\mathcal{G}$ independent of $\boldsymbol{\theta}$. In the following, we approximate the full expectation over $\mathbf{z}$ with a finite set of samples $\mathcal{Z} = \{\mathbf{z}_i\}$. Using the approximation to the true gradient given by $\mathbf{A}_{\mathbf{x}}$ and $\mathbf{b}_{\mathbf{x}}$, we can then easily derive a stochastic approximation of the gradient of the KL diver-

gence in Eq. (6) with respect to the parameters $\boldsymbol{\theta}$:

$$\nabla_{\boldsymbol{\theta}} \operatorname{KL}(q \,\|\, p)$$
$$\stackrel{(6)}{=} -\mathbb{E}_{\mathbf{z}\sim\mathcal{G}}\Big[\nabla_{\mathbf{x}} \log p\big(\mathbf{x}(\mathbf{z}) \,|\, \mathbf{y}\big) \cdot \nabla_{\boldsymbol{\theta}}\, \mathbf{x}(\mathbf{z})\Big] - \nabla_{\boldsymbol{\theta}}\, H(q) \tag{7a}$$

$$\approx -\frac{1}{|\mathcal{Z}|} \sum_{\mathbf{z}_i \in \mathcal{Z}} \nabla_{\mathbf{x}} \log p\big(\mathbf{x}(\mathbf{z}_i) \,|\, \mathbf{y}\big) \cdot \nabla_{\boldsymbol{\theta}}\, \mathbf{x}(\mathbf{z}_i) - \nabla_{\boldsymbol{\theta}}\, H(q) \tag{7b}$$

$$\stackrel{(4)}{\approx} \frac{1}{|\mathcal{Z}|} \sum_{\mathbf{z}_i \in \mathcal{Z}} \Big(\mathbf{A}_{\mathbf{x}}\big(\mathbf{x}(\mathbf{z}_i)\big)\mathbf{x}(\mathbf{z}_i) + \mathbf{b}_{\mathbf{x}}\big(\mathbf{x}(\mathbf{z}_i)\big)\Big) \cdot \nabla_{\boldsymbol{\theta}}\, \mathbf{x}(\mathbf{z}_i)$$

$$- \nabla_{\boldsymbol{\theta}}\, H(q) \tag{7c}$$
$$\equiv \bar{\nabla}_{\boldsymbol{\theta}} \operatorname{KL}(q \,\|\, p). \tag{7d}$$

**Gaussian mean field inference.** To illustrate the use of this approximation, we now apply the common naive mean-field framework [11, 21, 23] and assume that the variational distribution $q$ factorizes along all elements of $\mathbf{x} = (x_l)_l$ for $l = 1, \ldots, L$. Moreover, $q$ is modeled as an uncorrelated Gaussian distribution with $\boldsymbol{\theta} = \{\boldsymbol{\mu}, \boldsymbol{\sigma}\}$:

$$q(\mathbf{x}) = \prod_{l=1}^{L} \mathcal{N}(x_l \,|\, \mu_l, \sigma_l^2). \tag{8}$$

Following [21], $\mathbf{z}$ is thus chosen to be standard normally distributed, *i.e.* $\mathbf{z} \sim \mathcal{N}(0, \mathbf{I})$, and we set $\mathbf{x}(\mathbf{z}) = \mathbf{z} \cdot \boldsymbol{\sigma} + \boldsymbol{\mu}$ with element-wise operations.

For the case of a fully-factorized Gaussian $q$, it is now possible to express $\bar{\nabla}_{\boldsymbol{\theta}} \operatorname{KL}(q \,\|\, p)$ again in the form of a linearized gradient. To do this, we consider the individual parameter gradients w.r.t. $\boldsymbol{\mu}$ and $\boldsymbol{\sigma}$. For the gradient with respect to $\boldsymbol{\mu}$, we exploit that the entropy of a Gaussian distribution does not depend on its mean. Hence, we arrive at

$$\bar{\nabla}_{\boldsymbol{\mu}} \operatorname{KL}(q \,\|\, p)$$
$$= \frac{1}{|\mathcal{Z}|} \sum_{\mathbf{z}_i \in \mathcal{Z}} \Big(\mathbf{A}_{\mathbf{x}}\big(\mathbf{x}(\mathbf{z}_i)\big)\mathbf{x}(\mathbf{z}_i) + \mathbf{b}_{\mathbf{x}}(\mathbf{x}(\mathbf{z}_i))\Big) \cdot \nabla_{\boldsymbol{\mu}}\, \mathbf{x}(\mathbf{z}_i)$$
$$- \nabla_{\boldsymbol{\mu}}\, H(q) \tag{9a}$$
$$= \frac{1}{|\mathcal{Z}|} \sum_{\mathbf{z}_i \in \mathcal{Z}} \mathbf{A}_{\mathbf{x}}\big(\mathbf{x}(\mathbf{z}_i)\big)\big(\mathbf{z}_i \cdot \boldsymbol{\sigma} + \boldsymbol{\mu}\big) + \mathbf{b}_{\mathbf{x}}\big(\mathbf{x}(\mathbf{z}_i)\big) \tag{9b}$$
$$= \left[\frac{1}{|\mathcal{Z}|} \sum_{\mathbf{z}_i \in \mathcal{Z}} \mathbf{A}_{\mathbf{x}}\big(\mathbf{x}(\mathbf{z}_i)\big)\right] \boldsymbol{\mu}$$
$$+ \left[\frac{1}{|\mathcal{Z}|} \sum_{\mathbf{z}_i \in \mathcal{Z}} \mathbf{A}_{\mathbf{x}}\big(\mathbf{x}(\mathbf{z}_i)\big) \mathbf{D}(\mathbf{z}_i)\right] \boldsymbol{\sigma}$$
$$+ \left[\frac{1}{|\mathcal{Z}|} \sum_{\mathbf{z}_i \in \mathcal{Z}} \mathbf{b}_{\mathbf{x}}\big(\mathbf{x}(\mathbf{z}_i)\big)\right] \tag{9c}$$
$$\equiv \mathbf{A}_{\boldsymbol{\mu},\boldsymbol{\mu}}(\boldsymbol{\theta})\, \boldsymbol{\mu} + \mathbf{A}_{\boldsymbol{\mu},\boldsymbol{\sigma}}(\boldsymbol{\theta})\, \boldsymbol{\sigma} + \mathbf{b}_{\boldsymbol{\mu}}(\boldsymbol{\theta}), \tag{9d}$$

where $\mathbf{D}(\mathbf{z}_i)$ denotes a diagonal matrix comprised of the elements of $\mathbf{z}_i$. The gradient w.r.t. $\boldsymbol{\sigma}$ involves the derivative of the Gaussian entropy, *i.e.* $\nabla_{\boldsymbol{\sigma}} H(q) = \nabla_{\boldsymbol{\sigma}} (\log \boldsymbol{\sigma} + \text{const})$, which can be linearized in several ways. We opt for using the element-wise second-order Taylor expansion of the logarithm around the current estimate: $\boldsymbol{\sigma}^{(t)}$:

$$\log \boldsymbol{\sigma} \approx \log \boldsymbol{\sigma}^{(t)} + \frac{1}{\boldsymbol{\sigma}^{(t)}}\big(\boldsymbol{\sigma} - \boldsymbol{\sigma}^{(t)}\big) - \frac{1}{\big(\boldsymbol{\sigma}^{(t)}\big)^2}\big(\boldsymbol{\sigma} - \boldsymbol{\sigma}^{(t)}\big)^2 \tag{10a}$$

$$= \frac{1}{\boldsymbol{\sigma}^{(t)}} \boldsymbol{\sigma} - \frac{1}{\big(\boldsymbol{\sigma}^{(t)}\big)^2}\big(\boldsymbol{\sigma} - \boldsymbol{\sigma}^{(t)}\big)^2 + \text{const}. \tag{10b}$$

With that we can derive our stochastic approximation to the linearized gradient of the KL divergence w.r.t. $\boldsymbol{\sigma}$ as

$$\bar{\nabla}_{\boldsymbol{\sigma}} \operatorname{KL}(q \,\|\, p)$$
$$= \frac{1}{|\mathcal{Z}|} \sum_{\mathbf{z}_i \in \mathcal{Z}} \Big(\mathbf{A}_{\mathbf{x}}\big(\mathbf{x}(\mathbf{z}_i)\big)\mathbf{x}(\mathbf{z}_i) + \mathbf{b}_{\mathbf{x}}\big(\mathbf{x}(\mathbf{z}_i)\big)\Big) \cdot \nabla_{\boldsymbol{\sigma}}\mathbf{x}(\mathbf{z}_i)$$
$$- \nabla_{\boldsymbol{\sigma}}\, H(q) \tag{11a}$$
$$\approx \frac{1}{|\mathcal{Z}|} \sum_{\mathbf{z}_i \in \mathcal{Z}} \mathbf{D}(\mathbf{z}_i)\Big(\mathbf{A}_{\mathbf{x}}\big(\mathbf{x}(\mathbf{z}_i)\big)\big(\mathbf{z}_i \cdot \boldsymbol{\sigma} + \boldsymbol{\mu}\big) + \mathbf{b}_{\mathbf{x}}\big(\mathbf{x}(\mathbf{z}_i)\big)\Big)$$
$$- \frac{3}{\boldsymbol{\sigma}^{(t)}} + \frac{2}{\big(\boldsymbol{\sigma}^{(t)}\big)^2} \boldsymbol{\sigma} \tag{11b}$$
$$= \left[\frac{1}{|\mathcal{Z}|} \sum_{\mathbf{z}_i \in \mathcal{Z}} \mathbf{D}(\mathbf{z}_i)\mathbf{A}_{\mathbf{x}}\big(\mathbf{x}(\mathbf{z}_i)\big)\right]\boldsymbol{\mu}$$
$$+ \left[\frac{1}{|\mathcal{Z}|} \sum_{\mathbf{z}_i \in \mathcal{Z}} \mathbf{D}(\mathbf{z}_i)\mathbf{A}_{\mathbf{x}}\big(\mathbf{x}(\mathbf{z}_i)\big)\mathbf{D}(\mathbf{z}_i) + \frac{2}{\big(\boldsymbol{\sigma}^{(t)}\big)^2}\right]\boldsymbol{\sigma}$$
$$+ \left[\frac{1}{|\mathcal{Z}|} \sum_{\mathbf{z}_i \in \mathcal{Z}} \mathbf{z}_i \mathbf{b}_{\mathbf{x}}\big(\mathbf{x}(\mathbf{z}_i)\big) - \frac{3}{\boldsymbol{\sigma}^{(t)}}\right] \tag{11c}$$
$$\equiv \mathbf{A}_{\boldsymbol{\sigma},\boldsymbol{\mu}}(\boldsymbol{\theta})\, \boldsymbol{\mu} + \mathbf{A}_{\boldsymbol{\sigma},\boldsymbol{\sigma}}(\boldsymbol{\theta})\, \boldsymbol{\sigma} + \mathbf{b}_{\boldsymbol{\sigma}}(\boldsymbol{\theta}). \tag{11d}$$

From Eqs. (9d) and (11d), we now obtain an approximate linearized gradient of the KL divergence in Eq. (2) with respect to $\boldsymbol{\mu}$ and $\boldsymbol{\sigma}$. Following the GL procedure, the optimization proceeds by solving the linear system of equations

$$\boldsymbol{\theta}^{(t+1)} = -\mathbf{A}_{\boldsymbol{\theta}}\big(\boldsymbol{\theta}^{(t)}\big)^{-1} \mathbf{b}_{\boldsymbol{\theta}}\big(\boldsymbol{\theta}^{(t)}\big) \tag{12}$$

with

$$\mathbf{A}_{\boldsymbol{\theta}}(\boldsymbol{\theta}) = \begin{bmatrix} \mathbf{A}_{\boldsymbol{\mu},\boldsymbol{\mu}}(\boldsymbol{\theta}) & \mathbf{A}_{\boldsymbol{\mu},\boldsymbol{\sigma}}(\boldsymbol{\theta}) \\ \mathbf{A}_{\boldsymbol{\sigma},\boldsymbol{\mu}}(\boldsymbol{\theta}) & \mathbf{A}_{\boldsymbol{\sigma},\boldsymbol{\sigma}}(\boldsymbol{\theta}) \end{bmatrix}, \quad \mathbf{b}_{\boldsymbol{\theta}}(\boldsymbol{\theta}) = \begin{bmatrix} \mathbf{b}_{\boldsymbol{\mu}}(\boldsymbol{\theta}) \\ \mathbf{b}_{\boldsymbol{\sigma}}(\boldsymbol{\theta}) \end{bmatrix}. \tag{13}$$

Note that we can treat the underlying energy $E$ as a black box. The only interaction with $E$ is through its linearized gradient. Algorithm 1 summarizes our approach.

**Discussion.** Each gradient iteration in Eq. (12) can be interpreted as fitting a quadratic function to the Monte Carlo approximation of the KL divergence (Eq. 7b), such that the quadratic approximation and the KL divergence agree on their first-order derivatives at $\boldsymbol{\theta}^{(t)}$. This alone does not guarantee that the extremum $\boldsymbol{\theta}^{(t+1)}$ of the quadratic function is actually a minimum of the approximation. Hence, we now show that the Hessian of the quadratic approximation $\mathbf{A}_{\boldsymbol{\theta}}$ is positive semi-definite, thus ensuring that $\boldsymbol{\theta}^{(t+1)}$ minimizes the approximated KL divergence.

**Proposition 1.** $\mathbf{A}_{\boldsymbol{\theta}}(\boldsymbol{\theta}^{(t)})$ *is positive semi-definite, i.e.* $\boldsymbol{\theta}^T \mathbf{A}_{\boldsymbol{\theta}}(\boldsymbol{\theta}^{(t)})\boldsymbol{\theta} \geq 0, \forall \boldsymbol{\theta}, \boldsymbol{\theta}^{(t)} \in \mathbb{R}^{2L}$, *if the matrix* $\mathbf{A}_{\mathbf{x}}(\mathbf{x}(\mathbf{z}))$ *of the energy GL is positive semi-definite for all* $\mathbf{x}(\mathbf{z})$.

*Proof.* Let us first assume that we just draw a single sample $\mathbf{z}$. To simplify notation let $\mathbf{A}_{\mathbf{x}} \equiv \mathbf{A}_{\mathbf{x}}(\mathbf{x}(\mathbf{z}))$ and $\mathbf{A}_{\boldsymbol{\theta}} \equiv \mathbf{A}_{\boldsymbol{\theta}}(\boldsymbol{\theta}^{(t)})$. Now, for $\boldsymbol{\theta} = [\boldsymbol{\mu}, \boldsymbol{\sigma}]^T$ we have that

$$\boldsymbol{\theta}^T \mathbf{A}_{\boldsymbol{\theta}} \boldsymbol{\theta}$$
$$= \boldsymbol{\mu}^T \mathbf{A}_{\boldsymbol{\mu},\boldsymbol{\mu}} \boldsymbol{\mu} + \boldsymbol{\sigma}^T \mathbf{A}_{\boldsymbol{\sigma},\boldsymbol{\mu}} \boldsymbol{\mu} + \boldsymbol{\mu}^T \mathbf{A}_{\boldsymbol{\mu},\boldsymbol{\sigma}} \boldsymbol{\sigma} + \boldsymbol{\sigma}^T \mathbf{A}_{\boldsymbol{\sigma},\boldsymbol{\sigma}} \boldsymbol{\sigma} \tag{14a}$$
$$= \boldsymbol{\mu}^T \mathbf{A}_{\mathbf{x}} \boldsymbol{\mu} + \boldsymbol{\sigma}^T \mathbf{D}(\mathbf{z})^T \mathbf{A}_{\mathbf{x}} \boldsymbol{\mu} + \boldsymbol{\mu}^T \mathbf{A}_{\mathbf{x}} \mathbf{D}(\mathbf{z}) \boldsymbol{\sigma} \tag{14b}$$
$$+ \boldsymbol{\sigma}^T \mathbf{D}(\mathbf{z})^T \left( \mathbf{A}_{\mathbf{x}} + \mathbf{D}\left( 2/(\boldsymbol{\sigma}^{(i)})^2 \right) \right) \mathbf{D}(\mathbf{z}) \boldsymbol{\sigma}$$
$$= (\boldsymbol{\mu} + \mathbf{D}(\mathbf{z})\boldsymbol{\sigma})^T \mathbf{A}_{\mathbf{x}} (\boldsymbol{\mu} + \mathbf{D}(\mathbf{z})\boldsymbol{\sigma}) \tag{14c}$$
$$+ (\mathbf{D}(\mathbf{z})\boldsymbol{\sigma})^T \mathbf{D}\left( 2/(\boldsymbol{\sigma}^{(i)})^2 \right) (\mathbf{D}(\mathbf{z})\boldsymbol{\sigma})$$
$$\geq 0, \tag{14d}$$

where we inserted the definition of the individual matrices (Eqs. 9d and 11d). For the last step, we used our assumption that $\mathbf{A}_{\mathbf{x}}$ is positive semi-definite. The case of multiple samples $\mathbf{z}_i$ can be shown analogously by expanding each of the four terms in Eq. (14a) into a sum. □

To put the above proposition into perspective, we now give two mild conditions on the energy function such that the corresponding matrix $\mathbf{A}_{\mathbf{x}}$ is positive semi-definite.

**Proposition 2.** *An energy function can be linearized with a positive semi-definite matrix* $\mathbf{A}_{\mathbf{x}}$ *if it is composed of a sum of energy terms* $\rho_i(\mathbf{w}_i)$ *that fulfill the following conditions:*

1. *Each penalty function* $\rho_i(\cdot)$ *is symmetric and* $\rho_i'(\mathbf{w}_i) \geq 0$ *for all* $\mathbf{w}_i \geq 0$. ($\star$)

2. *Each penalty function* $\rho_i(\cdot)$ *is applied element-wise on* $\mathbf{w}_i$, *which is of the form* $\mathbf{w}_i = \mathbf{K}_i \mathbf{x} + \mathbf{g}_i(\mathbf{y})$, *with filter matrix* $\mathbf{K}_i$ *and function* $\mathbf{g}_i$ *not depending on* $\mathbf{x}$. ($\star\star$)

*Proof.* See supplemental material. □

The above assumptions of Proposition 2 are not very restrictive but met by many MRF/CRF potentials [6], including the smoothness term used in optical flow and Poisson-Gaussian denoising, as well as the data term of our flow

---

**Algorithm 1** Gaussian mean field inference with SVIGL

**Require:** $\boldsymbol{\theta}^{(0)}$: Initial variational parameters
$\mathbf{A}_{\mathbf{x}}, \mathbf{b}_{\mathbf{x}}$: Gradient linearization of the model energy
**for** $t = 0, \ldots, T-1$ **do**
  Generate samples $\mathbf{z}_i$
  $\mathbf{x}_i \leftarrow \boldsymbol{\sigma} \cdot \mathbf{z}_i + \boldsymbol{\mu}$
  Compute $\mathbf{A}_{\mathbf{x}}(\mathbf{x}_i)$ and $\mathbf{b}_{\mathbf{x}}(\mathbf{x}_i)$
  Compute $\mathbf{A}_{\boldsymbol{\theta}}(\boldsymbol{\theta}^{(t)})$ and $\mathbf{b}_{\boldsymbol{\theta}}(\boldsymbol{\theta}^{(t)})$ as in Eq. (13)
  $\boldsymbol{\theta}^{(t+1)} \leftarrow -\mathbf{A}_{\boldsymbol{\theta}}(\boldsymbol{\theta}^{(t)})^{-1} \mathbf{b}_{\boldsymbol{\theta}}(\boldsymbol{\theta}^{(t)})$
**end for**
**return** $\boldsymbol{\theta}^{(T)}$

---

energy, *c.f.* Sec. 5.1 and 5.2. Moreover, positive semi-definiteness of $\mathbf{A}_{\mathbf{x}}$ can also be shown for more complex energy formulations such as the data term of Poisson-Gaussian denoising used in our experiments.

**Implementation details.** Solving the linear system of equations of Eq. (12) exactly is too costly for many large-scale problems, which may involve millions of variables. Hence, we consistently apply 100 iterations of successive over-relaxation [48] with a relaxation factor of 1.95 and the current estimate $\boldsymbol{\theta}^{(t)}$ as initialization. We also experimented with a conjugate gradient optimizer, but found convergence to be too slow, probably due to the need of an effective preconditioner. One limitation of our method is that we cannot guarantee that $\boldsymbol{\sigma}$ stays positive after each optimization step. Therefore, we replace each new iterate $\boldsymbol{\sigma}^{(t+1)}$ with its absolute value. In practice, however, we found that usually the entropy term is enough to force $\boldsymbol{\sigma}$ to stay positive. Since the gradient of the KL divergence cannot be expressed conveniently as linear in $\log \boldsymbol{\sigma}$, we do not use the usual trick of optimizing for $\log \boldsymbol{\sigma}$ to directly enforce positivity of $\boldsymbol{\sigma}$.

## 5. Experiments

We now demonstrate that SVIGL provides a convenient and efficient way of obtaining accurate variational approximations for popular energy functions of diverse computer vision problems, yielding uncertainty estimates that correlate well with estimation errors. Specifically, we quantitatively evaluate on the tasks of optical flow estimation and Poisson-Gaussian denoising. We compare SVIGL against gradient-based optimization of the KL divergence with SGD as well as the Adam optimizer [20], the default choice in the popular Edward library [40]. To assess the quality of the obtained approximate posterior, we evaluate the KL divergence $\text{KL}(q \| p)$ as well as application specific performance metrics. We always report KL divergences approximated by sampling (*c.f.* Eq. 6) and up to the unknown, but constant log partition function $\log Z(\mathbf{y})$.

We conduct several experiments for each application. We begin by evaluating the robustness of Adam (in the con-text of stochastic variational inference) and SVIGL w.r.t. to their parameters. We first vary the step size $\alpha$ of Adam while using $|\mathcal{Z}| = 50$ samples per iteration to approximate the KL divergence gradient. Next, we use the best step size and vary the size of the sample set $|\mathcal{Z}|$ for both Adam and SVIGL. For a sample set size of 50, 25, and 12, we set the number of iterations to 100, 200, and 400 for SVIGL and 1000, 2000, and 4000 for Adam, respectively. For SGD, we similarly tune the hyperparameters and find that 4000 iterations with 12 samples and an initial step size of $10^{-6}$, which is cut after each third of iterations by a factor of ten, works best for both applications. We compare the best configurations of SVIGL and SVI with SGD and Adam to a Laplace approximation and MAP estimation baselines. Runtimes refer to an Intel Xeon E5-2650v4, 2.2 GHz, 12 cores. We furthermore show qualitative results for 3D surface reconstruction to demonstrate the benefit of SVIGL for non-vision applications.

### 5.1. Optical flow

We first apply SVIGL to estimate an optical flow field $\mathbf{x}$, describing the motion between images $\mathbf{y} = \{I_1, I_2\}$. We use the EpicFlow energy of [32] to induce a Gibbs distribution akin to Eq. (1). Its likelihood encourages the flow to be consistent with the images and is based on a gradient consistency assumption, whereas the prior assumes small flow gradients over a 4-neighborhood, *i.e.*

$$E(\mathbf{x}, \mathbf{y}) = \lambda_\text{D} \sum_{l=1}^{L} \rho_\text{D}\left(\left\|(\nabla \tilde{I}_2(\mathbf{x}) - \nabla I_1)_l\right\|_2\right) \quad (15)$$
$$+ \lambda_\text{S} \sum_{j=1}^{J} \sum_{l=1}^{L} \rho_\text{S}\left(\left\|(\mathbf{f}_j * \mathbf{x})_l\right\|_2\right).$$

Here, $\nabla I_1$ denotes the spatial derivatives of $I_1$, $\tilde{I}_2(\mathbf{x})$ is the second image warped by $\mathbf{x}$, and $\mathbf{f}_1, \ldots, \mathbf{f}_J$ represent (derivative) filters. Functions $\rho_\text{D}$ and $\rho_\text{S}$ are robust penalty functions weighted with parameters $\lambda_\text{D}, \lambda_\text{S}$. Following standard practice, we linearize the likelihood around the current flow.

**Setup.** As in [45], we initialize our estimates with sparse FlowFields matches [1], densified with the EpicFlow interpolation [32]. Variances are initialized as $\sigma = 10^{-3}$. We use generalized Charbonnier penalties [2] and obtain their parameters as well as the ratio $\lambda_\text{D}/\lambda_\text{S}$ through Bayesian optimization [37]. To that end, we evaluate the average endpoint error (AEPE) of MAP estimates on a subset of Sintel train [9]. The absolute scale of $\lambda_\text{D}$ and $\lambda_\text{S}$ is subsequently calibrated such that the AEPE of the SVIGL estimates remains comparable to the MAP estimates on the training set.

**Results.** We conduct experiments on a validation set of 104 images randomly chosen from Sintel training (excluding images used for parameter optimization). We first motivate the use of gradient linearization by comparing the results of MAP estimation performed with up to 200 iterations of L-BFGS to 20 iterations of GL. The results averaged over the validation set are depicted in Fig. 2. We observe a significantly faster minimization of the energy using GL, which highlights its benefits for highly non-linear objectives.

We now compare SVIGL to SVI with Adam. In order to keep the runtime of Adam manageable, we perform the evaluation on manually cropped patches of size $100 \times 100$. In a first setting, we vary the step size $\alpha$ of Adam using 1000 iterations for Adam and 100 iterations for SVIGL. In Fig. 3a, we evaluate the KL divergence plotted against the runtime. SVIGL reduces the KL divergence two orders of magnitude faster than Adam on this challenging energy function. Moreover, the optimization by Adam is highly dependent on the chosen step size; too small or too large a value may equally lead to slow convergence. In contrast, SVIGL does not require the selection of a step size. For the following experiments we fix the step size for Adam to $\alpha = 0.005$. Now, we vary the number of samples and iterations as described above. The results are shown in Fig. 3b. Again, SVIGL attains a significantly better variational approximation than SVI with Adam for all examined settings.

Table 1 summarizes the KL divergence and the average runtime for the best settings of Adam ($\alpha = 0.01$, $|\mathcal{Z}| = 12$), SGD, and SVIGL ($|\mathcal{Z}| = 12$). In a similar runtime, SVIGL achieves a significantly lower KL divergence than SVI with Adam or SGD. We additionally evaluate the diagonal Laplace approximation around the MAP estimates using the Hessian of the linearized energy. SVIGL shows a moderate improvement over the Laplace approximation. However, the Laplace method requires second-order derivatives, which are tedious and error-prone to derive. Moreover, the Laplace approximation does not lead to consistently good results, *c.f.* Sec. 5.2.

Finally, we evaluate SVIGL on the *full-size* images of Sintel test. Since SVI with Adam is too slow, we only compare to MAP baselines with 200 iterations of L-BFGS and 20 iterations of GL, respectively. For SVIGL we use 50 samples and also 20 iterations. Both SVIGL and GL yield an AEPE of 5.74 and therefore outperform the L-BFGS baseline with an AEPE of 5.81.

**Interpretation.** The interdependent updates of SVIGL (Eq. 12) causes information to flow between all variables while a regular gradient step propagates information in a local spatial neighborhood only. We attribute the observed performance gain of GL and SVIGL over gradient-based methods at least partly to this global update.

Table 1. Unnormalized KL divergence and average runtime on $100 \times 100$ crops from a Sintel validation set.

| Method | KL[$*10^7$] | runtime [s] |
| --- | --- | --- |
| Initialization | 5.13 | – |
| GL + Laplace | 3.83 | – |
| SVI + SGD | 4.45 | **551** |
| SVI + Adam | 4.24 | 1148 |
| SVIGL *(ours)* | **3.78** | 584 |

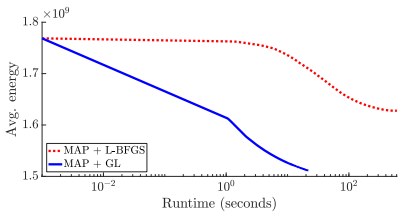
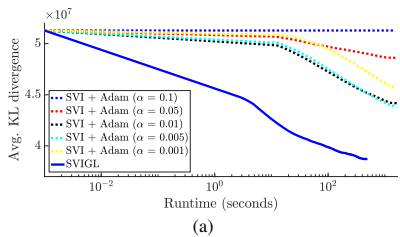
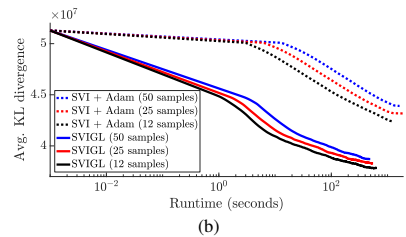

Figure 2. Optical flow energy *vs.* runtime for MAP estimation with L-BFGS and GL. Values averaged over the validation dataset. GL is clearly superior to standard L-BFGS.

Figure 3. Unnormalized KL divergence *vs.* runtime for SVIGL and SVI with Adam on optical flow with different step sizes (a) and different numbers of samples and iterations (b). Values averaged over the validation set.

**Uncertainty estimates.** Finally, we assess the quality of the per-pixel uncertainty estimates. To this end, we compare to the recent strong baseline ProbFlowFields [45]. Specifically, we apply SVIGL to update the continuous variables of their energy formulation; see supplemental material for further implementational details. Table 2 shows the metrics introduced in [45], averaged over the full-size images of our validation set. The uncertainty estimates obtained by SVIGL are competitive with the ones of ProbFlowFields. More importantly and unlike [45], the application of SVIGL does not require the tedious derivation of update equations. Example flow fields and the inferred uncertainty maps are shown in Fig. 1.

### 5.2. Poisson-Gaussian denoising

We next apply SVIGL to the problem of removing Poisson-Gaussian noise [15]. Here, it is assumed that image noise comes mainly from two sources that inherently affect any camera sensor. First, the Poissonian arrival process of photons hitting the pixels, and second an additive Gaussian component arising from noise in the electronics of the sensor. The Poisson distribution can be well approximated by a Gaussian [15], giving rise to a Gaussian likelihood with intensity dependent variance, *i.e.*

$$\mathbf{y}_l \sim \mathcal{N}\big(\mathbf{x}_l, \sigma(\mathbf{x}_l)^2\big) \text{ with } \sigma(\mathbf{x}_l)^2 = \beta_1 \mathbf{x}_l + \beta_2, \quad (16)$$

where the noise distribution is specified by the parameters $\beta_1$ and $\beta_2$. We specifically set $\beta_1 = 0.05$ and $\beta_2 = 0.0001$ in order to simulate strong noise (Poisson rate 20). Combining this likelihood with a 4-connected pairwise MRF with generalized Charbonnier potentials [2] as image prior leads to the energy

$$E(\mathbf{x}, \mathbf{y}) = \frac{\lambda_\mathrm{D}}{2} \sum_{l=1}^{L} \frac{(\mathbf{x}_l - \mathbf{y}_l)^2}{\sigma(\mathbf{x}_l)^2} + \lambda_\mathrm{S} \sum_{j=1}^{J} \sum_{l=1}^{L} \rho_S\Big(\big(\mathbf{f}_j * \mathbf{x}\big)_l\Big), \quad (17)$$

where the $\mathbf{f}_j$ denote horizontal and vertical image derivative filters. The temperature is subsumed by the weights $\lambda_\mathrm{D}, \lambda_\mathrm{S}$.

Table 2. AEPE, area under curve (AUC) of the sparsification plots, and Spearman's rank correlation coefficient for SVIGL and ProbFlowFields on our validation set, *c.f.* [45] for further details. †Difference in AEPE is caused by one outlier image pair.

| Method | AEPE | AUC | CC |
|---|---|---|---|
| ProbFlowFields [45] | 3.13 | 0.40 | 0.56 |
| SVIGL *(ours)* | 3.21† | 0.42 | 0.50 |

**Setup.** We select the relative importance of $\lambda_\mathrm{D}$ and $\lambda_\mathrm{S}$ as well as the exponent of the robust penalty through Bayesian optimization [37]. To this end, we optimize the peak-signal-to-noise ratio (PSNR) after 20 steps of GL on a set of 100 images from the BSDS training set [26]. We then calibrate the posterior for VI by determining the absolute scale of the weights on the training set. To synthesize noisy images for parameter tuning and testing, we apply Poisson-Gaussian noise to clean ground truth images. Afterwards, we rescale the intensities such that the ground truth lies in $[0, 1]$ and clip the noisy image to that range. For test time inference, we initialize $\boldsymbol{\mu}$ with the noisy image and $\boldsymbol{\sigma}$ as $10^{-3}$.

**Results.** In Fig. 4 we plot the unnormalized KL divergence against runtime for SVIGL and SVI with Adam, using varying step sizes for Adam and varying sizes of the sample set $\mathcal{Z}$ for both methods. It becomes apparent that the performance of Adam highly depends on these two parameters.

Table 3. Unnormalized KL divergences, PSNR values, and SSIM [44] for SVIGL and baseline methods in denoising.

| Method | KL [$*10^6$] | PSNR [dB] | SSIM |
|---|---|---|---|
| Initialization | 1.95 | 17.29 | 0.287 |
| GL + Laplace | 1.57 | 24.71 | 0.662 |
| SVI + SGD | 1.23 | 19.49 | 0.384 |
| SVI + Adam | 0.98 | 24.70 | 0.680 |
| SVIGL *(ours)* | **0.97** | **24.77** | **0.693** |
| MAP + L-BFGS | – | 23.17 | 0.605 |
| MAP + GL | – | 24.71 | 0.662 |

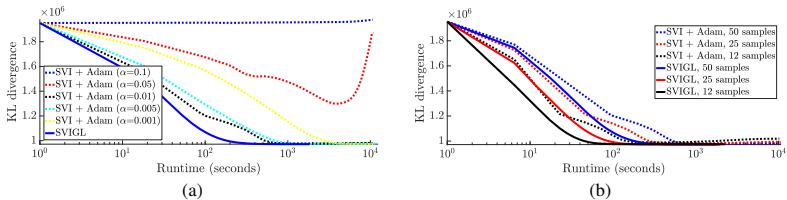
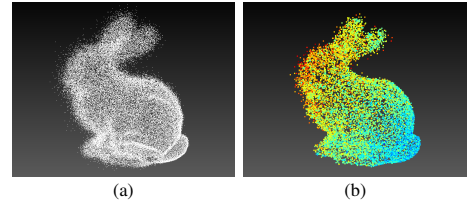

Figure 4. Runtime *vs.* unnormalized KL divergence for denoising with SVIGL and SVI with Adam with different stepsize parameters $\alpha$ (a) and varying sizes of the sample set $|\mathcal{Z}|$ (b). Values averaged over the BSDS test set.

Figure 5. Noisy input point cloud (a) and smoothed point cloud (b); colors indicate posterior uncertainty (blue – low, red – high).

Too small a step size slows down convergence, while setting it too high leads to a KL divergence inferior to the initialization. In contrast, SVIGL does not require setting a step size and converges faster than Adam with the best step size $\alpha = 0.01$. For instance, SVIGL reaches the same KL divergence as Adam in only $1/5$ of the time. When looking at the size of the sample set, we note that smaller sample sets speed up each iteration and hence lead to faster progress of the optimization. However, the solution found by Adam deteriorates after a certain number of iterations with smaller sample set sizes, while SVIGL is not affected by this issue. In summary, SVIGL yields faster convergence while being robust to the setting of nuisance parameters.

The converged solutions are evaluated in Table 3. SIVGL ($|\mathcal{Z}| = 50$) not only converges significantly faster than Adam ($\alpha = 0.01$, $|\mathcal{Z}| = 50$), but obtains even slightly improved solutions. SGD performs significantly worse than SVIGL and Adam. A Laplace approximation around the mode obtained with 100 iterations of GL provides a poor fit to the denoising posterior since the dependence of the variances $\sigma(\mathbf{x}_l)$ on the noise-free intensities $\mathbf{x}_l$ results in a skewed distribution. Furthermore, we see that SVIGL obtains a better solution in terms of the standard image quality metrics PSNR and SSIM [44] than the MAP estimation baselines obtained with GL and L-BFGS, *e.g.* +1.6 dB in PSNR compared to L-BFGS. In the supplemental material we show denoised images obtained by SVIGL along with their uncertainty estimates.

### 5.3. 3D surface reconstruction

In order to demonstrate that SVIGL is not limited to low-level problems in computer vision, we apply it to the task of reconstructing a smooth point cloud from noisy input data. Specifically, we use the energy of [25] given as

$$E(X, P, C) = \sum_{i=1}^{|X|} \sum_{j=1}^{|P|} \|x_i - p_j\| \cdot h(\|c_i - p_j\|) \quad (18)$$
$$- \sum_{i=1}^{|X|} \sum_{i'=1}^{|C|} \lambda_i \|x_i - c_{i'}\| \cdot h(\|c_i - c_{i'}\|).$$

Here, $p_j \in P$ denote the noisy input points; the current and the new estimate of the smoothed points are given by $c_i \in C$ and $x_i \in X$, respectively. The contribution of each term is weighted by a Gaussian kernel $h(\cdot)$. Following Lipman *et al.* [25], we use this energy in a fixed point scheme, *i.e.*

$$X_{t+1} = \arg\min_X E(X, P, X_t), \quad (19)$$

where $X_0$ is an $L_2$ projection of the input points. The supplemental material describes the setup in more detail.

In order to exemplify the use of SVIGL for 3D surface reconstruction, we synthesize a noisy input point cloud of the Stanford bunny by adding noise on the positions of reference points. The noise strength gradually increases from tail to face. Figure 5 shows both the noisy input point cloud as well as the variational approximation from SVIGL with color coded uncertainty $\boldsymbol{\sigma}$. It is apparent that the uncertainty increases with input noise strength, thus reflecting the difficulty of the reconstruction task. Moreover, at points further away from the true surface, the uncertainty is generally higher, *c.f.* the outliers at the ears.

## 6. Conclusion

Motivated by the success of gradient linearization techniques for MAP estimation in highly multimodal posteriors, we proposed to combine the benefits of gradient linearization with stochastic variational inference. As a result we obtain SVIGL, an easy-to-use variational inference scheme that only requires access to a gradient linearization of the posterior energy and allows to simply repurpose well-proven energy minimization schemes. We applied SVIGL to optical flow estimation as well as Poisson-Gaussian denoising and demonstrated its significantly faster convergence compared to standard stochastic variational inference. Moreover, we showed that the optimization accuracy of SVIGL is robust to the choice of parameters. The inferred uncertainty estimates are competitive with state-of-the-art but can be obtained without tedious derivations of update equations. Finally, we demonstrate that SVIGL is not restricted to dense 2D prediction tasks by applying it successfully to the task of 3D surface reconstruction.

**Acknowledgments.** The research leading to these results has received funding from the European Research Council under the European Union's Seventh Framework Programme (FP/2007–2013)/ERC Grant agreement No. 307942.

# Stochastic Variational Inference with Gradient Linearization
## – Supplemental Material –

Tobias Plötz*     Anne S. Wannenwetsch*     Stefan Roth
Department of Computer Science, TU Darmstadt

**Preface.** In this supplemental material, we show that SVIGL can be interpreted as a gradient descent approach using a special preconditioner and provide the gradient linearization for the optical flow and Poisson-Gaussian energies used in the main manuscript. Furthermore, we provide a proof for Proposition 2, show the hyperparameter evaluation for SVI with SGD, and give additional details of our optical flow experiment in Table 2. Finally, we show some exemplary results of Poisson-Gaussian denoising and provide details on the experiment on 3D surface reconstruction.

## A. SVIGL as Preconditioned Gradient Descent

Here, we show that an update step of SVIGL as given in Eq. (12) can be interpreted as one iteration of preconditioned gradient descent. To simplify notation let $\mathbf{A}_{\boldsymbol{\theta}} \equiv \mathbf{A}_{\boldsymbol{\theta}}(\boldsymbol{\theta}^{(t)})$ and $\mathbf{b}_{\boldsymbol{\theta}} \equiv \mathbf{b}_{\boldsymbol{\theta}}(\boldsymbol{\theta}^{(t)})$. Following, *e.g.* [29], we have

$$\boldsymbol{\theta}^{(t+1)} = -\mathbf{A}_{\boldsymbol{\theta}}^{-1}\mathbf{b}_{\boldsymbol{\theta}} \qquad (20a)$$

$$= \boldsymbol{\theta}^{(t)} - \mathbf{A}_{\boldsymbol{\theta}}^{-1}\mathbf{b}_{\boldsymbol{\theta}} - \boldsymbol{\theta}^{(t)} \qquad (20b)$$

$$= \boldsymbol{\theta}^{(t)} - \mathbf{A}_{\boldsymbol{\theta}}^{-1}\left(\mathbf{b}_{\boldsymbol{\theta}} + \mathbf{A}_{\boldsymbol{\theta}}\boldsymbol{\theta}^{(t)}\right) \qquad (20c)$$

$$= \boldsymbol{\theta}^{(t)} - \mathbf{A}_{\boldsymbol{\theta}}^{-1}\nabla_{\boldsymbol{\theta}}\operatorname{KL}(q\,\|\,p). \qquad (20d)$$

Therefore, SVIGL performs gradient descent with preconditioner $P = \mathbf{A}_{\boldsymbol{\theta}}^{-1}$. This interpretation also allows to introduce a step size parameter $\alpha$ to SVIGL

$$\boldsymbol{\theta}^{(t+1)} = \boldsymbol{\theta}^{(t)} - \alpha\mathbf{A}_{\boldsymbol{\theta}}^{-1}\nabla_{\boldsymbol{\theta}}\operatorname{KL}(q\,\|\,p) \qquad (21a)$$

$$= \boldsymbol{\theta}^{(t)} - \alpha\mathbf{A}_{\boldsymbol{\theta}}^{-1}\left(\mathbf{b}_{\boldsymbol{\theta}} + \mathbf{A}_{\boldsymbol{\theta}}\boldsymbol{\theta}^{(t)}\right) \qquad (21b)$$

$$= (1-\alpha)\boldsymbol{\theta}^{(t)} + \alpha\hat{\boldsymbol{\theta}}^{(t+1)}, \qquad (21c)$$

with $\hat{\boldsymbol{\theta}}^{(t+1)} = -\mathbf{A}_{\boldsymbol{\theta}}^{-1}\mathbf{b}_{\boldsymbol{\theta}}$ denoting the SVIGL estimate as given in Eq. (12). In practice, our experiments have shown that the performance of SVIGL is not sensitive to the choice of the step size parameter. We thus simply set $\alpha = 1$.

*Authors contributed equally

## B. Linearized Gradients

In the following, we show how linearized gradients can be obtained for the presented applications of SVIGL in optical flow estimation and Poisson-Gaussian denoising. For other applications, including many models in computer vision, it is possible to derive parameters $\mathbf{A}_{\boldsymbol{\theta}}$ and $\mathbf{b}_{\boldsymbol{\theta}}$ in a similar fashion.

### B.1. Optical flow

Here, we show the derivation of a linearized gradient for a simple optical flow energy using the brightness constancy assumption, *i.e.*

$$E(\mathbf{x},\mathbf{y}) = \lambda_{\mathrm{D}}\sum_{l=1}^{L}\rho_{\mathrm{D}}\left(I_{t,l} + \begin{pmatrix}I_{x,l}\\I_{y,l}\end{pmatrix}^{\mathrm{T}}\left(\mathbf{x}_l - \mathbf{x}_l^0\right)\right)$$
$$+ \lambda_{\mathrm{S}}\sum_{j=1}^{J}\sum_{l=1}^{L}\rho_{\mathrm{S}}\left(\left\|(\mathbf{f}_j \ast \mathbf{x})_l\right\|_2\right) \qquad (22a)$$

$$= \lambda_{\mathrm{D}}E_{\mathrm{D}}(\mathbf{x},\mathbf{y}) + \lambda_{\mathrm{S}}E_{\mathrm{S}}(\mathbf{x}), \qquad (22b)$$

with $I_{t,l} = I_2\left(l + \mathbf{x}_l^0\right) - I_1(l)$, $\begin{pmatrix}I_{x,l}\\I_{y,l}\end{pmatrix} = \nabla I_2\left(l + \mathbf{x}_l^0\right)$, and $\mathbf{x}_l^0$ denoting the point of approximation of the Taylor linearization. The derivations for the EpicFlow energy function in Eq. (15) are more tedious, but can be done analogously.

**Data term.** In a first step, we derive the linearized gradient for the data energy term. Here, it holds that

$$\nabla_{\mathbf{x}_l}E_{\mathrm{D}}(\mathbf{x},\mathbf{y}) = \nabla_{\mathbf{x}_l}\rho_{\mathrm{D}}\left(I_{t,l} + \begin{pmatrix}I_{x,l}\\I_{y,l}\end{pmatrix}^{\mathrm{T}}\left(\mathbf{x}_l - \mathbf{x}_l^0\right)\right) \qquad (23a)$$

$$= \rho_{\mathrm{D}}'\left(I_{t,l} + \begin{pmatrix}I_{x,l}\\I_{y,l}\end{pmatrix}^{\mathrm{T}}\left(\mathbf{x}_l - \mathbf{x}_l^0\right)\right)\cdot\begin{pmatrix}I_{x,l}\\I_{y,l}\end{pmatrix}. \qquad (23b)$$

The derivative of the generalized Charbonnier [2] used for

$\rho_D(\cdot)$ can be written as:

$$\rho'_D(x) = \frac{x}{c^2}\left(\frac{(x/c)^2}{\max(1, 2-a)} + 1\right)^{(a/2-1)} \quad (24a)$$

$$\equiv \tilde{\rho}_D(x)\, x. \quad (24b)$$

Using Eqs. (23b) and (24b), we have

$$\nabla_{\mathbf{x}_l} E_D(\mathbf{x}, \mathbf{y}) =$$
$$= \tilde{\rho}_D\left(I_{t,l} + \begin{pmatrix}I_{x,l}\\I_{y,l}\end{pmatrix}^T (\mathbf{x}_l - \mathbf{x}_l^0)\right)$$
$$\cdot \left(\begin{pmatrix}I_{x,l}I_{t,l}\\I_{y,l}I_{t,l}\end{pmatrix} + \begin{pmatrix}I_{x,l}^2 & I_{x,l}I_{y,l}\\I_{x,l}I_{y,l} & I_{y,l}^2\end{pmatrix}(\mathbf{x}_l - \mathbf{x}_l^0)\right). \quad (25)$$

The last identity (Eq. 25) allows us to easily identify a linearized form of the gradient of the data term as

$$\nabla_{\mathbf{x}} E_D(\mathbf{x}, \mathbf{y}) = \mathbf{A}_{\mathbf{x}}^D(\mathbf{x})\mathbf{x} + \mathbf{b}_{\mathbf{x}}^D(\mathbf{x}), \quad (26)$$

with

$$\mathbf{A}_{\mathbf{x}}^D(\mathbf{x}) = \begin{pmatrix}\mathbf{D}(\tilde{\rho}_D \cdot I_x^2) & \mathbf{D}(\tilde{\rho}_D \cdot I_x I_y)\\ \mathbf{D}(\tilde{\rho}_D \cdot I_x I_y) & \mathbf{D}(\tilde{\rho}_D \cdot I_y^2)\end{pmatrix} \quad (27)$$

and

$$\mathbf{b}_{\mathbf{x}}^D(\mathbf{x}) = \begin{pmatrix}\mathbf{D}(\tilde{\rho}_D \cdot I_x I_t)\mathbf{1}\\ \mathbf{D}(\tilde{\rho}_D \cdot I_y I_t)\mathbf{1}\end{pmatrix} - \mathbf{A}_{\mathbf{x}}(\mathbf{x})\mathbf{x}^0. \quad (28)$$

Here, $\mathbf{x} = \left(x_1^{(1)}, \ldots, x_L^{(1)}, x_1^{(2)}, \ldots, x_L^{(2)}\right)^T$ denotes the stacked vector of all horizontal and vertical flow components. $\mathbf{D}(\cdot)$ turns the argument vector into a diagonal matrix (short for diag$\{\cdot\}$), and the product is applied element-wise.

**Smoothness term.** For the smoothness term let us first express the convolution $\mathbf{f}_j * \mathbf{x}$ as a matrix-vector product $\mathbf{F}_j \cdot \mathbf{x}$, with $\mathbf{F}_j$ denoting the convolution matrix corresponding to $\mathbf{f}_j$ and $\mathbf{x}$ the vectorized flow as before. With that, the gradient of the smoothness term $E_S$ can be written as:

$$\nabla_{\mathbf{x}} E_S(\mathbf{x}) = \nabla_{\mathbf{x}} \sum_{j=1}^{J}\sum_{l=1}^{L} \rho_S\big((\mathbf{F}_j \mathbf{x})_l\big) \quad (29a)$$

$$= \sum_{j=1}^{J} \mathbf{F}_j^T \rho_S'(\mathbf{F}_j \mathbf{x}). \quad (29b)$$

Using the derivative $\rho_S'$ as given in Eq. (24b), we obtain

$$\sum_{j=1}^{J} \mathbf{F}_j^T \rho_S'(\mathbf{F}_j \mathbf{x}) = \sum_{j=1}^{J} \mathbf{F}_j^T \, \mathbf{D}\big(\tilde{\rho}_S(\mathbf{F}_j \mathbf{x})\big) \mathbf{F}_j \mathbf{x} \quad (30a)$$

$$= \left(\sum_{j=1}^{J} \mathbf{F}_j^T \, \mathbf{D}\big(\tilde{\rho}_S(\mathbf{F}_j \mathbf{x})\big) \mathbf{F}_j\right) \mathbf{x} \quad (30b)$$

$$\equiv \mathbf{A}_{\mathbf{x}}^S(\mathbf{x})\mathbf{x}. \quad (30c)$$

**Complete linearized gradient.** We now summarize the results of Eqs. (27), (28), and (30c) to obtain the linearized gradient as

$$\nabla_{\mathbf{x}} E(\mathbf{x}, \mathbf{y}) = \lambda_D \nabla_{\mathbf{x}} E_D(\mathbf{x}, \mathbf{y}) + \lambda_S \nabla_{\mathbf{x}} E_S(\mathbf{x}) \quad (31a)$$

$$= \big(\lambda_D \mathbf{A}_{\mathbf{x}}^D(\mathbf{x}) + \lambda_S \mathbf{A}_{\mathbf{x}}^S(\mathbf{x})\big)\mathbf{x} + \lambda_D \mathbf{b}_{\mathbf{x}}^D \quad (31b)$$

$$\equiv \mathbf{A}_{\mathbf{x}}(\mathbf{x})\mathbf{x} + \mathbf{b}_{\mathbf{x}}. \quad (31c)$$

### B.2. Poisson-Gaussian denoising

Let us first recap the energy function for Poisson-Gaussian denoising:

$$E(\mathbf{x}, \mathbf{y}) = \frac{\lambda_D}{2} \sum_{l=1}^{L} \frac{(\mathbf{x}_l - \mathbf{y}_l)^2}{\sigma(\mathbf{x}_l)^2} \quad (32a)$$
$$+ \lambda_S \sum_{j=1}^{J}\sum_{l=1}^{L} \rho_S\big((\mathbf{f}_j * \mathbf{x})_l\big),$$

$$= \lambda_D E_D(\mathbf{x}, \mathbf{y}) + \lambda_S E_S(\mathbf{x}), \quad (32b)$$

where

$$\sigma(\mathbf{x}_l)^2 = \beta_1 \mathbf{x}_l + \beta_2. \quad (33)$$

We will derive the linearized gradients for the data term $E_D$ and the smoothness term $E_S$ separately.

**Data term.** The gradient of the data term is given as

$$\nabla_{\mathbf{x}} E_D(\mathbf{x}, \mathbf{y})$$
$$= \frac{(\mathbf{x} - \mathbf{y})}{\sigma(\mathbf{x})^2} - \frac{\beta_1(\mathbf{x} - \mathbf{y})^2}{2\sigma(\mathbf{x})^4} \quad (34a)$$
$$= \frac{\mathbf{x}}{\sigma(\mathbf{x})^2} - \frac{\mathbf{y}}{\sigma(\mathbf{x})^2} - \frac{\beta_1 \mathbf{x}^2}{2\sigma(\mathbf{x})^4} + \frac{\beta_1 \mathbf{x}\mathbf{y}}{\sigma(\mathbf{x})^4} - \frac{\beta_1 \mathbf{y}^2}{2\sigma(\mathbf{x})^4} \quad (34b)$$
$$= \mathbf{x}\left(\frac{1}{\sigma(\mathbf{x})^2} - \frac{\beta_1 \mathbf{x}}{2\sigma(\mathbf{x})^4} + \frac{\beta_1 \mathbf{y}}{\sigma(\mathbf{x})^4}\right)$$
$$- \left(\frac{\mathbf{y}}{\sigma(\mathbf{x})^2} + \frac{\beta_1 \mathbf{y}^2}{2\sigma(\mathbf{x})^4}\right), \quad (34c)$$

where all operations are element-wise. The linearized gradient of the data term can then be obtained as

$$\mathbf{A}_{\mathbf{x}}^{\text{D}}(\mathbf{x}) = \mathbf{D}\left(\frac{1}{\sigma(\mathbf{x})^2} - \frac{\beta_1 \mathbf{x}}{2\sigma(\mathbf{x})^4} + \frac{\beta_1 \mathbf{y}}{\sigma(\mathbf{x})^4}\right) \quad (35)$$

$$\mathbf{b}_{\mathbf{x}}^{\text{D}}(\mathbf{x}) = -\left(\frac{\mathbf{y}}{\sigma(\mathbf{x})^2} + \frac{\beta_1 \mathbf{y}^2}{2\sigma(\mathbf{x})^4}\right). \quad (36)$$

**Smoothness term.** For the smoothness term we can reuse the linearized gradient derived in Eq. (30c).

**Complete linearized gradient.** We can now put the results of Eqs. (30c), (35), and (36) together to obtain a linearized gradient of the energy for Poisson-Gaussian denoising, *c.f.* Eqs. (31a) – (31c).

## C. Proof Proposition 2

In this section, we provide a proof for Proposition 2 of the main paper.

**Proposition 2.** *An energy function can be linearized with a positive semi-definite matrix $\mathbf{A}_\mathbf{x}$ if it is composed of a sum of energy terms $\rho_i(\mathbf{w}_i)$ that fulfill the following conditions:*

1. *Each penalty function $\rho_i(\cdot)$ is symmetric and $\rho_i'(\mathbf{w}_i) \geq 0$ for all $\mathbf{w}_i \geq 0$.* ($\star$)

2. *Each penalty function $\rho_i(\cdot)$ is applied element-wise on $\mathbf{w}_i$, which is of the form $\mathbf{w}_i = \mathbf{K}_i \mathbf{x} + \mathbf{g}_i(\mathbf{y})$, with filter matrix $\mathbf{K}_i$ and $\mathbf{g}_i$ not depending on $\mathbf{x}$.* ($\star\star$)

*Proof.* From assuming a symmetric $\rho_i(\cdot)$ in ($\star$), it follows that $\rho_i'(\cdot)$ is point symmetric. Due to $\rho_i'(\mathbf{w}_i) \geq 0$ for all $\mathbf{w}_i \geq 0$ we then find that $\rho_i'(\mathbf{w}_i)$ can be written as

$$\rho_i'(\mathbf{w}_i) \equiv \tilde{\rho}_i(\mathbf{w}_i) \cdot \mathbf{w}_i \quad \text{with a} \quad \tilde{\rho}_i(\mathbf{w}_i) \geq 0. \quad (37)$$

For an energy term as described in ($\star\star$), the gradient w.r.t. $\mathbf{x}$ is given as

$$\nabla_\mathbf{x} \rho_i(\mathbf{w}_i) = \mathbf{K}_i^\text{T} \cdot \mathbf{C}_i \cdot (\mathbf{K}_i \cdot \mathbf{x} + \mathbf{g}_i(\mathbf{y})), \quad (38)$$

$$\text{with} \quad \mathbf{C}_i = \mathbf{D}\big(\tilde{\rho}_i\left(\mathbf{K}_i \cdot \mathbf{x} + \mathbf{g}_i(\mathbf{y})\right)\big). \quad (39)$$

A linearization can then be obtained using

$$\mathbf{A}_\mathbf{x}^i = \mathbf{K}_i^\text{T} \cdot \mathbf{C}_i \cdot \mathbf{K}_i, \qquad \mathbf{b}_\mathbf{x}^i = \mathbf{K}_i^\text{T} \cdot \mathbf{C}_i \cdot \mathbf{g}_i(\mathbf{y}). \quad (40)$$

Since $\mathbf{C}_i$ is a diagonal matrix of non-negative elements (Eq. 37), $\mathbf{A}_\mathbf{x}^i$ is positive semi-definite as

$$\mathbf{x}^\text{T} \mathbf{A}_\mathbf{x}^i \mathbf{x} = \mathbf{x}^\text{T} \mathbf{K}_i^\text{T} \mathbf{C}_i \mathbf{K}_i \mathbf{x} = \mathbf{v}^\text{T} \mathbf{C}_i \mathbf{v} \geq 0. \quad (41)$$

As the sum of positive semi-definite matrices is positive semi-definite, a matrix $\mathbf{A}_\mathbf{x}$ composed of energy terms that fulfill ($\star$) and ($\star\star$) is positive semi-definite. □

## D. Hyperparameters for SGD

In the following, we aim to find optimal hyperparameters for the SVI baseline based on SGD. For all experiments we select an initial step size $\alpha_0$, which is cut after each third of iterations by a factor of ten. An evaluation of the unnormalized KL divergence for optical flow plotted against the runtime for different initial step sizes $\alpha_0$ of SGD is shown in Fig. 6a. Here, the KL divergence deteriorates severely using SGD with a step size larger than $10^{-6}$. For smaller step sizes, SVI with SGD shows a slow convergence such that we set $\alpha_0 = 10^{-6}$.

Following the same procedure, we perform several experiments for Poisson-Gaussian denoising and evaluate different settings for the initial step size parameter $\alpha_0$ of SGD in Fig. 7a. Again, an initial step size $\alpha_0 = 10^{-6}$ proves to be most effective. Smaller step sizes converge too slowly, while SGD with bigger step size values converges faster but to a worse local optimum. For an initial step size of $\alpha_0 = 10^{-5}$ optimization diverges immediately.

Applying SVI with SGD, we observe in both applications a faster convergence of the KL divergence with a smaller sample size, but a larger number of iterations, *c.f.* Figs. 6b and 7b. We therefore choose $|\mathcal{Z}| = 12$ with 4000 iterations of SGD for the experiments in the main paper.

## E. Comparison with ProbFlowFields

In Table 2 of the main paper we evaluate the quality of the posterior variances obtained with SVIGL. Here, we follow Wannenwetsch *et al.* [45] and derive an uncertainty measure by computing the marginal entropy of the flow at every pixel. To have a fair comparison with [45], we use the same EpicFlow [32] energy formulation with learned Gaussian scale mixture penalty functions and explicit indicator variables for their mixture components. Since SVIGL is designed for variational inference in distributions with continuous random variables, we alternate closed-form updates of the latent indicator variables with SVIGL updates for the continuous flow variables. For the discrete update, we approximate the tedious analytical expectation values over the flow variables with a Monte-Carlo estimator (*c.f.* Eq. 7b). This effectively reduces the optimization w.r.t. the indicator variables to an independent update – thus maintaining the ease of use of SVIGL. Weighting parameters $\lambda_\text{D}$ and $\lambda_\text{S}$ are determined on a training set with Bayesian optimization [37] using the F1-score as described in [45].

## F. Results of Poisson-Gaussian Denoising

Fig. 8 shows some example results of SVIGL applied to Poisson-Gaussian denoising on the BSDS dataset. High uncertainties can be observed especially on object boundaries. Due to the high amount of noise, a strong smoothness term maximizes the PSNR on the training set. Therefore, the denoised images tend to be rather smooth in general.

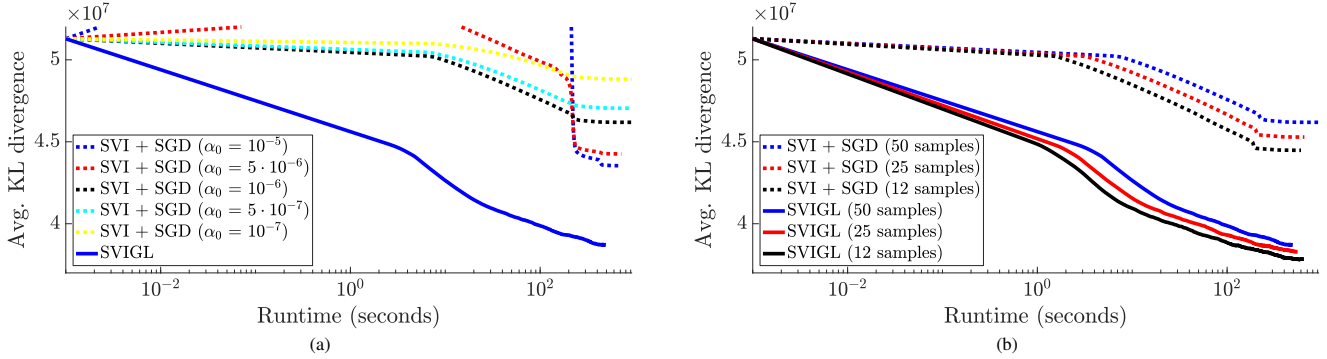

Figure 6. Unnormalized KL divergence *vs.* runtime for optical flow with SVIGL and SVI with SGD with different step sizes (a) and different numbers of samples and iterations (b). Values averaged on the validation set.

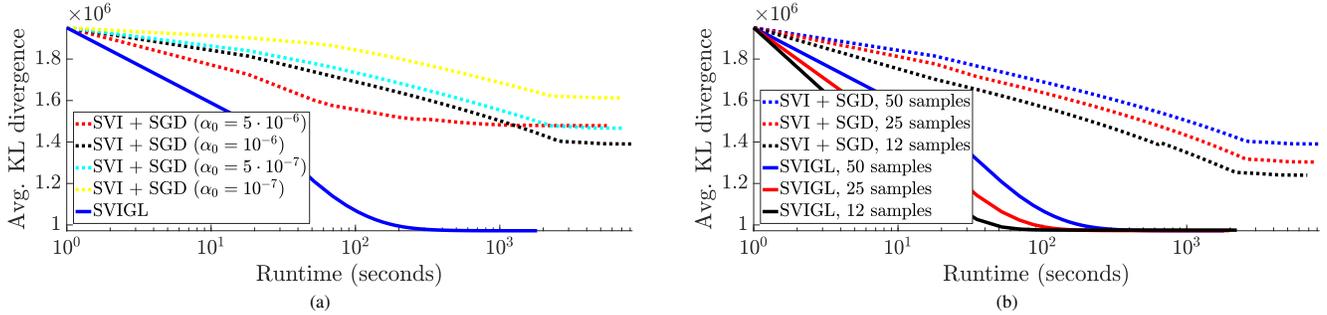

Figure 7. Unnormalized KL divergence *vs.* runtime for denoising with SVIGL and SVI with SGD with different step sizes (a) and with different numbers of samples and iterations (b). Values averaged over the BSDS test set.

## G. Results on Sintel Test

As described in Sec. 5.1, we apply SVIGL as well as two MAP baselines on the full-sized Sintel test images in order to evaluate their performance. Figure 9 shows a screenshot of the private Sintel benchmark table with results for both methods. SVIGL outperforms the underlying FlowFields method [1] as well as the L-BFGS baseline and shows an AEPE result on par with the corresponding MAP estimate using GL. Moreover, SVIGL estimates are competitive with the finetuned version of FlowNet2 [49], *i.e.* the state-of-the-art baseline for optical flow prediction with convolutional neural networks.

## H. 3D Surface Reconstruction

We now give more details on the application of SVIGL to 3D surface reconstruction. First, we restate the energy of Lipman *et al.* [25], which is given by

$$E(X, P, C) = \sum_{i=1}^{|X|} \sum_{j=1}^{|P|} \|x_i - p_j\| \cdot h(\|c_i - p_j\|)$$
$$- \sum_{i=1}^{|X|} \sum_{i'=1}^{|C|} \lambda_i \|x_i - c_{i'}\| \cdot h(\|c_i - c_{i'}\|). \quad (42)$$

Here, $p_j \in P$ denote the noisy input points, $c_i \in C$ are the current estimates of the smoothed points, and $x_i \in X$ the new estimates of the smoothed points. While the first part of the energy forces the new estimates to be close to the input points, the second term pushes the reconstructed points apart by penalizing points in $X$ that are too close to points in $C$. The contribution of each term is weighted by the Gaussian kernel $h(\cdot)$.

A closed-form solution to minimizing the above energy is given in [25]. This solution is then used in a fixed point scheme as

$$X_{t+1} = \arg\min_{X} E(X, P, X_t), \quad (43)$$

where $X_0$ is initialized as a $L_2$ projection of the input points.

In a variational inference setting, closed-form updates are no longer possible due to introducing the additional vari-ance variables $\sigma$ of the variational posterior. Hence, we employ SVIGL updates instead. To be able to apply SVIGL, we require a linearization of the energy gradient. The specific form of the energy in Eq. (19) allows for a diagonal linearization:

$$\nabla_{x_i} E(X, P, C) = \sum_{j \in J} (x_i - p_j) \frac{h(\|c_i - p_j\|)}{\|x_i - p_j\|}$$
$$- \sum_{i' \in I} (x_i - c_{i'}) \frac{h(\|c_i - c_{i'}\|)}{\|x_i - c_{i'}\|}. \quad (44)$$

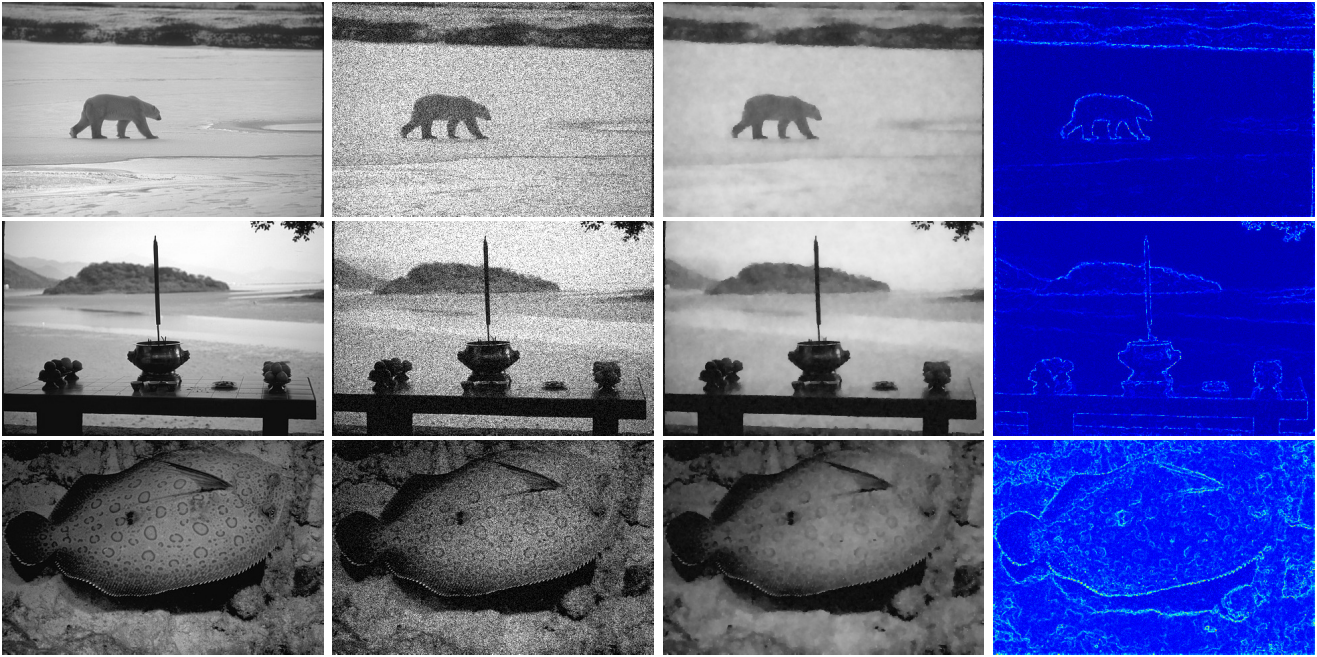

Figure 8. Examples of ground truth (left), noisy images (second column), estimated clean images (third column), and uncertainty estimates (right) from SVIGL on the BSDS test set.

|  | EPE all | EPE matched | EPE unmatched | d0-10 | d10-60 | d60-140 | s0-10 | s10-40 | s40+ |
|---|---|---|---|---|---|---|---|---|---|
| GroundTruth [1] | 0.000 | 0.000 | 0.000 | 0.000 | 0.000 | 0.000 | 0.000 | 0.000 | 0.000 |
| ⋮ |  |  |  | ⋮ |  |  |  |  | ⋮ |
| MAP_GL [25] | 5.735 | 2.524 | 31.896 | 4.789 | 2.126 | 1.619 | 1.107 | 3.659 | 33.708 |
| FlowNet2-ft-sintel [26] | 5.739 | 2.752 | 30.108 | 4.818 | 2.557 | 1.735 | 0.959 | 3.228 | 35.538 |
| SVIGL [27] | 5.740 | 2.526 | 31.924 | 4.792 | 2.128 | 1.619 | 1.107 | 3.661 | 33.740 |
| MAP_LBFGS [28] | 5.805 | 2.626 | 31.710 | 4.890 | 2.233 | 1.709 | 1.206 | 3.803 | 33.456 |
| FlowFields [29] | 5.810 | 2.621 | 31.799 | 4.851 | 2.232 | 1.682 | 1.157 | 3.739 | 33.890 |

Figure 9. Screenshot of the private Sintel benchmark table (final) with results for SVIGL, MAP + GL, MAP + L-BFGS, and the original FlowFields approach [1] (status as of March 2018).

In total, we run 10 iterations of Eq. (43). In each iteration, we compute a single SVIGL update with a sample set size of $|\mathcal{Z}| = 5$.